\documentclass{article}

\usepackage[final]{corl_2019} % Uncomment for the camera-ready ``final'' version

\usepackage{amsmath,amsfonts,bm}

\def\eqref#1{equation~\ref{#1}}
\def\1{\bm{1}}

\DeclareMathAlphabet{\mathsfit}{\encodingdefault}{\sfdefault}{m}{sl}
\SetMathAlphabet{\mathsfit}{bold}{\encodingdefault}{\sfdefault}{bx}{n}

\DeclareMathOperator*{\argmin}{arg\,min}

\usepackage{hyperref}
\usepackage{url}

\usepackage{subfigure}
\usepackage{amsmath,amssymb,amsfonts}
\usepackage{bbm}
\usepackage{graphicx}	% includegrapics
\usepackage{xcolor}		% textcolor
\usepackage{booktabs}	% professional-quality tables
\usepackage{algorithm}
\usepackage[noend]{algpseudocode}
\usepackage{enumitem}
\usepackage{textcomp}
\usepackage[font=small]{caption}
\usepackage{hyperref}   % hyperlinks
\usepackage{url}        % simple URL typesetting
\usepackage{nicefrac}   % compact symbols for 1/2, etc.

\usepackage[page]{appendix}
\usepackage{microtype}
\usepackage{titlesec}

\setlist{nosep}

\titlespacing{\chapter}{0pt}{*.1}{*.1}
\titlespacing{\section}{0pt}{*.1}{*.1}
\titlespacing{\subsection}{0pt}{*.1}{*.1}
\titlespacing{\subsubsection}{0pt}{*.1}{*.1}
\titlespacing{\paragraph}{0pt}{*.1}{*1}
\titlespacing{\subparagraph}{0pt}{*.1}{*1}

\def\BState{\State\hskip-\ALG@thistlm}

\usepackage{url}

\newcommand{\cyl}[1]{\textcolor{black}{#1}}

\usepackage{lipsum}

\title{Adversarial Active Exploration for \\Inverse Dynamics Model Learning}

\author{
    Zhang-Wei Hong
\and
    \textbf{Tsu-Jui Fu}
\and
    \textbf{Tzu-Yun Shann}
\and
    \textbf{Yi-Hsiang Chang}
\and
    \textbf{Chun-Yi Lee} \\
    \textit{\{williamd4112, rayfu1996ozig, arielshann, shawn420, cylee\}@gapp.nthu.edu.tw} \\
Elsa Lab, Department of Computer Science \\ National Tsing Hua University, Hsinchu, Taiwan
}

\begin{document}
\maketitle

%===============================================================================
\vspace{-3ex}
\begin{abstract}
We present an adversarial active exploration for inverse dynamics model learning, a simple yet effective learning scheme that incentivizes exploration in an environment without any human intervention. Our framework consists of a  deep reinforcement learning (DRL) agent and an inverse dynamics model contesting with each other. The former collects training samples for the latter, with an objective to maximize the error of the latter. The latter is trained with samples collected by the former, and generates rewards for the former when it fails to predict the actual action taken by the former. In such a competitive setting, the DRL agent learns to generate samples that the inverse dynamics model fails to predict correctly, while the inverse dynamics model learns to adapt to the challenging samples. We further propose a reward structure that ensures the DRL agent to collect only moderately hard samples but not overly hard ones that prevent the inverse model from predicting effectively. We evaluate the effectiveness of our method on several robotic arm and hand manipulation tasks against multiple baseline models. Experimental results show that our method is comparable to those directly trained with expert demonstrations, and superior to the other baselines even without any human priors. 
\end{abstract}
\section{Introduction}

%\blfootnote{Demonstration videos: \url{http://bit.ly/2OpMeaE}}
%\TODO{Strengthen the motivation of using DNN}
% Problem Formulation
% Inverse model is useful in robotic control, but has some drawbacks in training data collection
Over the past decade, inverse dynamics models have shown considerable successes in robotic control~\cite{nguyen2010using,meier2016towards,rueckert2017learning,nguyen2011model} and even in humanoid robots~\cite{kanoun2011planning}. The main objective of an inverse dynamics model is to predict the control action (e.g., torque) between two states.  With such a model, it becomes practically feasible to carry out complex control behaviors through inferring a series of control actions given a pre-defined trajectory of states.  An inverse dynamics model is usually trained from streaming of data (i.e., states and actions) collected at online execution time, as it is extremely difficult to analytically specify the inverse dynamics model for a complicated robot.  Traditionally, a branch of research directions attempt to employ gaussian process (GP) to approximate the inverse dynamics model.  However, the computational complexity of such methods is intractable in high-dimensional state space, thus prohibiting their further usage from more sophisticated robotic control applications.%\blfootnote{Source code: \url{github.com/williamd4112/corl-adversarial-active-exploration}}

%Over the past decade, inverse dynamics models have shown considerable successes in robotic control~\cite{nguyen2010using,meier2016towards,rueckert2017learning,nguyen2011model} and even in humanoid robots~\cite{kanoun2011planning}. \zwh{An inverse dynamics model predicts the causal control action (e.g., torque) between two states. Roboticists realize the complex control behaviors through inferring a series of control actions of a given trajectory using an inverse dynamics model. An inverse dynamics model is usually learned from streaming of data at online execution time since it is difficult to analytically specify the inverse dynamics model in a complicated robot. Traditionally, local models like gaussian process (GP) that allow online learning are used to estimate the model. However, the computational complexity of local models is intractable in high-dimensional state space, thus prohibiting applications in more complicated anthropomorphic robots.}

% Recent works suggest using neural network inv model, but since online learning a neural network is infeasible, only batch training is feasible. Batch training would require tremendous amount of data, thus posing a challenge on efficient dataset collection.

In the light of the above constraint, researchers in recent years turn into developing inverse dynamics models based on deep neural networks (DNNs) in the hope to effectively cope with high-dimensional state space. \citet{rueckert2017learning} trains an inverse dynamics model using recurrent neural networks (RNNs), and demonstrates better prediction accuracy and computational efficiency than GP-based methods. \citet{pathak2018zeroshot,nair2017combining}, and~\citet{agrawal2016learning} further show that DNNs are able to learn an image-based inverse dynamics model for robotic arm manipulation tasks.  
% Camera-ready
In spite of their promising results on challenging tasks, however, this line of works demand tremendous amount of data for training DNNs, posing a considerable challenge on data-efficiency.

% Original
% In spite of their superior performance to the traditional GP-based methods, however, this line of works demand tremendous amount of data for training DNNs, posing a considerable challenge on data-efficiency.

% \zwh{Recent works train inverse dynamics models using neural networks in the hope to cope with high-dimensional data. ~\citet{rueckert2017learning} trains an inverse dynamics model using recurrent neural networks, showing better predictive accuracy and computational efficiency than GP. ~\citet{pathak2018zeroshot,nair2017combining,agrawal2016learning} further show that deep neural networks can learn an image-based inverse dynamics model for robotic arm manipulation. In spite of better performance and computational efficiency than GP, this line of works demand on tremendous amount of data for training neural networks, which poses a challenge on data-efficiency.}

As data efficiency is crucial for robot learning in practical applications, an efficient data acquisition methodology is critically essential for inverse dynamics model learning. Training data for an inverse dynamics model typically come from interactions with an environment. Previous approaches, however, usually perform such interactions with the environments in inefficient manners. {For instance, \citet{agrawal2016learning} and~\citet{nair2017combining} employ an agent to take random actions in an environment to collect training data for their inverse dynamics models.  Random actions, nevertheless, are less likely to result in effective exploration behaviors, and may thus lead to a lack of comprehensiveness in the data samples collected by the agent.  \citet{nair2017combining} attempts to deal with the above problem by adding human bias in its data sampling process, which is specifically tailored to certain pre-defined robot configurations.  Despite their promising results, tailored sampling ranges require significant amount of human effort.  A more general exploration strategy called curiosity-driven exploration was later proposed in~\cite{pathak2017curiosity}.  While this approach is effective in exploring novel states in an environment, their exploration behavior is solely driven by the prediction errors of the forward dynamics models.  Thus, the approach is unsuitable and not specifically tailored for inverse dynamics model learning.
%\zwh{While this method is effective in exploring novel states in the environment, their exploration is driven by the prediction errors of the forward dynamics models and thus cannot facilitate inverse dynamics model learning effectively}.
%While this method is effective in exploring novel states, it focuses only on training the forward dynamics model in its framework, and is not applicable to its inverse dynamics model.
}

In order to deal with the above issues, in this paper we propose a straightforward and efficient active data acquisition method, called \textbf{adversarial active exploration}, which motivates exploration of an environment in a self-supervised manner (i.e., without human intervention).  Inspired by~\citet{Pinto2017AdversarialRL,Shioya2018adversarialRL,sukhbaatar2018intrinsic}, we implement the proposed method by jointly training a deep reinforcement learning (DRL) agent and an inverse dynamics model competing with each other.  The former explores the environment to collect training data for the latter, and receives rewards from the latter if the data samples are considered difficult.  The latter is trained with the data collected by the former, and only generates rewards when it fails to predict the true actions performed by the former.  In such an adversarial setting, the DRL agent is rewarded only for the failure of the inverse dynamics model.  Therefore, the DRL agent learns to sample hard examples to maximize the chances to fail the inverse dynamics model.  On the other hand, the inverse dynamics model learns to be robust to the hard examples collected by the DRL agent by minimizing the probability of failures.  As a result, as the inverse dynamics model becomes stronger, the DRL agent is also incentivized to search for harder examples to obtain rewards.  Overly hard examples, however, may lead to biased exploration and cause instability of the learning process.  In order to stabilize the learning curve of the inverse dynamics model, we further propose a reward structure such that the DRL agent are encouraged to explore moderately hard examples for the inverse dynamics model, but refraining from too difficult ones for the latter to learn.  
The self-regulating feedback structure between the DRL agent and the inverse dynamics model enables them to automatically construct a curriculum for exploration.
%In this paper, we propose an adversarial exploration strategy to improve autonomous data collection. To collect data, a reinforcement learning agent receives the loss of the inverse dynamics model as rewards, encouraging the agent to explore novel states in the environment. As the total reward is maximized when the agent takes actions that \emph{adversarially challenge the developing dynamics model}, we therefore call this method an \textbf{adversarial exploration strategy}. We further present a number of training techniques to stabilize the learning progress. Compared to previous relevant works, we show that our proposed method not only accelerates data collection, but also collects a more diverse set of training samples, which makes our inverse dynamics model learn faster and better.

% Experiments
We perform extensive experiments to validate our method on multiple robotic arm and hand manipulation task environments simulated by the MuJoCo physics engine~\citep{Todorov2012mujoco}.
%, including \textit{FetchReach}, \textit{FetchPush}, \textit{FetchPickAndPlace}, \textit{FetchSlide}, and \textit{HandReach}. 
These environments are intentionally selected by us for evaluating the performance of inverse dynamics model, as each of them allows only a very limited set of chained actions to transition the robotic arms and hands to target states.  We examine the effectiveness of our method by comparing it against a number of data collection schemes.  The experimental results show that our method is more effective and data-efficient than the other schemes, as well as in environments with high-dimensional action spaces.  We also demonstrate that in most of the cases the performance of the inverse dynamics model trained by our method is comparable to that directly trained with expert trajectories.  The above observations suggest that our method is superior to the other data collection schemes even in the absence of human priors.  To justify each of our design decisions, we provide a comprehensive set of ablative analysis and discuss their implications.
%To demonstrate the effectiveness of our method, we evaluate our model in various robotic tasks in an environment powered by the MuJoCo physics engine, and compare our method against a number of baseline models. The experimental results show that our method is more effective and data-efficient than the baseline models in both low- and high-dimensional state spaces. We also find that our method outperforms the inverse dynamics model trained with human demonstrations, suggesting that our model is superior to the baselines even in the absence of human priors. We further evaluate our method in the environment with high-dimensional action spaces, and show that our model is significantly more effective than the baseline models.
% Contributions
The contributions of this work are summarized as the following:
\begin{itemize}[itemsep=.25ex]
\item {We introduce an adversarial active exploration method, which leverages a DRL agent to actively collect informative data samples for training an inverse dynamics model effectively.}
\item We employ a competitive scheme for the DRL agent and the inverse dynamics model, enabling them to automatically construct a curriculum for efficient data acquisition.
\item We introduce a reward structure for the proposed scheme to stabilize the training process.
\item {We demonstrate the effectiveness of the proposed method and compare it with a number of baseline data acquisition approaches for multiple robotic arm and hand manipulation tasks.}
\item We validate that our method is generalizable to tasks with high-dimensional action spaces.
\end{itemize}
The remainder of this paper is organized as follows. Section~\ref{sec::related_works} discusses the related works. Section~\ref{sec::backgrd} introduces background material necessary for understanding the contents of this paper. Section~\ref{sec::method} describes the proposed adversarial active exploration method in detail. Section~\ref{sec::exp} reports the experimental results, and provides an in-depth ablative analysis of our method. Section~\ref{sec::conclusion} concludes.
\section{Related Works}
\label{sec::related_works}
{This section reviews the related works of the areas: (i) active learning and (ii) exploration in DRL.}

Active learning~\cite{settles2009active} is an efficient way to query human for ground truth data labels when a large amount of unlabeled data are accessible while human labelling is expensive. Traditional works search for samples most uncertain for the model~\cite{lewis1994sequential,yang2015multi} or data that has the highest disagreement among an ensemble of models~\cite{seung1992query} {to query}. A recent work~\cite{fang2017learning} further trains {an DRL}~\cite{sutton2018reinforcement} agent to select informative training samples for a model, where unlabeled data are sequentially presented to the agent. {The method proposed in this paper differs from the above techniques in that it actively searches and collects data samples from an environment, rather than discriminate samples from a static dataset.}

Exploration in DRL refers to the process of searching for a better control policy to solve tasks via interactions with the environment. Apart from the naive random exploration approaches~\cite{sutton2000policy,watkins1992q}, a plethora of advanced exploration techniques~\cite{pathak2017curiosity,ostrovski2017count,stadie2015incentivizing,laversanne2018curiosity,colas2019curious,baranes2013active} has been proposed in the past few years to enhance the performance of DRL policies. The purpose of these works differs from ours in that our method aims to collect beneficial samples for learning an inverse dynamics model rather than f a policy. \cyl{In addition, some of these approaches are unable to be applied to our problem setting due to the absence of the goal space (i.e., representation of the task objectives) assumption in inverse dynamics model learning~\cite{laversanne2018curiosity,colas2019curious,baranes2013active}.
% Camera ready
A curiosity-driven DRL exploration technique which maximizes the prediction errors of the forward dynamics model~\cite{pathak2017curiosity} has recently been adapted to inverse dynamics model learning~\cite{pathak2018zeroshot}.} 
% Original
% {Moreover, a  curiosity-driven exploration technique~\cite{pathak2017curiosity,laversanne2018curiosity} has recently been applied to inverse dynamics model learning~\cite{pathak2018zeroshot} as well.} 
This approach encourages a DRL agent to visit novel states that a forward dynamics model is unable to accurately predict. However, novel states for a forward model do not necessarily reflect the novel states required for training an inverse dynamics model. {A comparison of our method and a data acquisition approach based on~\cite{pathak2017curiosity} is presented in Section~\ref{sec::exp}.}
% Original
% Curiosity-driven approaches ~\cite{pathak2017curiosity,laversanne2018curiosity,colas2019curious} basically encourage a DRL agent to visit novel states that a forward dynamics model is unable to accurately predict. However, novel states for a forward model do not necessarily reflect the novel states required for training an inverse dynamics model. {A comparison of our method and a data acquisition approach based on~\cite{pathak2017curiosity} is presented in Section~\ref{sec::exp}.}

\section{Background}
\label{sec::backgrd}
%In this section, we briefly review the concepts of DRL and inverse dynamics model for robotic control.

\subsection{Deep Reinforcement Learning}
\label{subsec::drl}
DRL methods train an agent to interact with an environment $\mathcal{E}$. At each timestep $t$, the agent receives an state $x_t \in \mathcal{X}$, where $\mathcal{X}$ is the state space of $\mathcal{E}$.  It then takes an action $a_t$ from the action space $\mathcal{A}$ based on its current policy $\pi$, receives a reward $r_t = r(x_t, a_t)$ (where $r$ is a reward function encoded with specific task objectives), and transitions to the next state $x^{\prime}$.  The policy $\pi:\mathcal{X} \times \mathcal{A} \mapsto [0, 1]$ is a conditional probability distribution parameterized by a deep neural network with parameters $\theta$.   The goal of the agent is to learn a $\pi$ that maximizes the discounted sum of rewards
$G_t=\sum^{T}_{\tau = t}\gamma^{\tau-t}r(x_{\tau},a_{\tau})$, where $t$ is the current timestep, $\gamma\in{(0, 1]}$ the discount factor, and $T$ the horizon. Policy gradient methods~\citep{sutton2000policy,williams1992simple} are a class of DRL techniques that directly optimize $\theta$ in the direction of $\nabla_\theta \mathbb{E}_{a_t \sim \pi} \big[\text{log} \pi(a_t|x_t, \theta) G_t$ \big].  In this paper, we employ a more recent family of policy gradient methods, called proximal policy optimization (PPO)~\citep{schulman2017proximal}. PPO is superior to {the vanilla policy gradient methods~\citep{williams1992simple} in terms of performance and sample efficiency.}

\subsection{Inverse Dynamics Model for Robotic Control}
\label{subsec::inv_model}
An inverse dynamics model $I$ takes as input a pair of states $(x, x^\prime)$, and predicts the action $\hat{a}$ required to reach the next state $x^\prime$ from the current state $x$. In this paper, we focus on parametric inverse dynamics models which can be formally express as: $\hat{a} = I(x, x^\prime|\theta_{I}),$
where ($x, x^\prime$) {can come from sensor measurements or a robot's states in the configuration space}, and $\theta_I$ represents the trainable parameters of $I$.  During the training phase, $\theta_I$ is iteratively updated to minimize the loss function $L_I$. The optimal $\theta_I$ is represented as: ${\theta_I}^* = \argmin_{\theta_I} L_I(a, \hat{a} | \theta_I) ~\forall x, x^\prime \in \mathcal{X}, a \in \mathcal{A},$  
where $a$ is the ground truth action.  Once {a sufficiently acceptable} ${\theta_I}^*$ is obtained, it can then be used to derive the control action sequence $\{\hat{a}_0, \hat{a}_1,\cdots, \hat{a}_{T-1}\}$ in a closed-loop manner, and therefore realize an arbitrarily given trajectory $\{\hat{x}_1,\cdots, \hat{x}_T\}$.  The derivation procedure is formulated as: $\hat{a}_t = I( x_t, \hat{x}_{t+1} | \theta_I) ~\forall t$, where $x_t$ and $\hat{x}_t$ denote the actual and the desired states, respectively.

\section{Methodology}
\label{sec::method}
    
In this section, we first describe the proposed adversarial active exploration. Then, we explain the training methodology in detail. Finally, we discuss a technique for stabilizing the training process.

\subsection{Adversarial Active Exploration}
\begin{figure}[t]
\vspace{-8ex}
\centering
\includegraphics[width=0.8\linewidth]{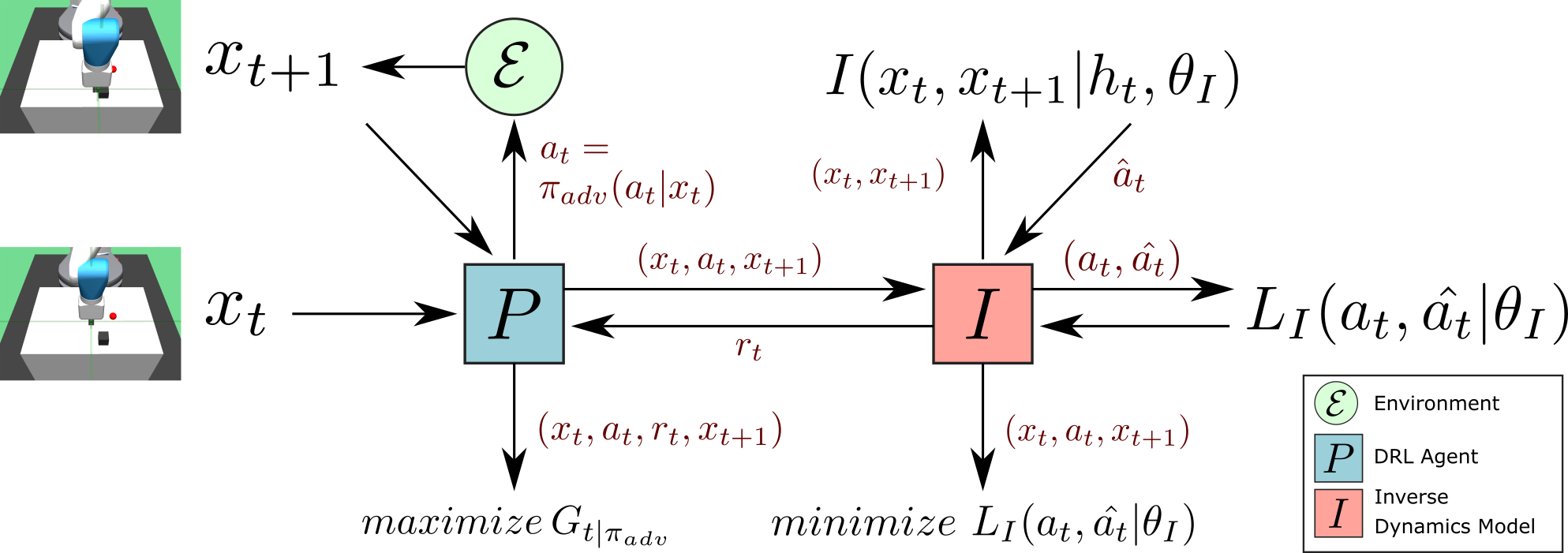}
\caption{Framework of adversarial active exploration.}
\label{figure::datacollection}
\end{figure}

Fig.~\ref{figure::datacollection} shows a framework that illustrates the proposed adversarial active exploration, which includes a DRL agent $P$ and an inverse dynamics model $I$. Assume that $\Phi_\pi:\{x_0, a_0, x_1, a_1\cdots,x_T\}$ is the sequence of states and actions generated by $P$ as it explores $\mathcal{E}$ using a policy $\pi$. At each timestep $t$, $P$ collects a 3-tuple training sample $(x_t, a_t, x_{t+1})$ for $I$, while $I$ predicts an action $\hat{a_t}$ and generates a reward $r_t$ for $P$. $I$ is modified to recurrently encode the information of its past states by including an additional hidden vector $h_t$, as suggested in~\cite{rueckert2017learning,pathak2018zeroshot}.  The inverse dynamics model $I$ is thus given by:
\begin{equation}
\label{eq::a_t}
\begin{aligned}
\footnotesize
   \hat{a_t} &= I(x_t, x_{t+1}|h_t, \theta_{I}) \\h_t &= f(h_{t-1},x_t),
\end{aligned}
\end{equation}
where $f(\cdot)$ denotes the recurrent function. Similar to \cite{rueckert2017learning,pathak2018zeroshot}, in this work we approximate ${\theta_I}$ in an offline batch training manner.  ${\theta_I}$ is iteratively updated to minimize $L_I$, formulated as the following:
\begin{equation}
\footnotesize
    \label{eq::inv_loss}
    \mathop{\text{minimize}}_{\theta_I} ~ \mathbb{E}_{(x_t, a_t, x_{t+1}) \sim U({Z}_I)} \big[ L_I(a_t, \hat{a}_t | \theta_I) \big],
\end{equation}
where $U({Z}_I)$ is an uniform distribution over the buffer $Z_I$ storing the 3-tuple data samples collected by $P$. We employ $\beta ||a_t -\hat{a_t}||^2$ (where $\beta$ is a scaling factor) as the loss function $L_I$, since we only consider continuous control domains in this paper. The loss function can be replaced with a cross-entropy loss for discrete control tasks. We directly use the loss function $L_I$ as the reward $r_t$ for $P$, where $r_t$ can be expressed as the following:
\begin{equation}
\label{eq::inv_reward}
\footnotesize
	r_t = r(x_t, a_t, x_{t+1}, {h_t}) =  L_I(a_t, \hat{a_t}| \theta_{I})= \beta  ||a_t -I(x_t, x_{t+1}|h_t, \theta_{I})||^2.
\end{equation}
Our method targets at improving both the quality and efficiency of the data collection process performed by $P$ as well as the performance of $I$ via collecting difficult and non-trivial training samples.  Therefore, the goal of the proposed framework is twofold.  First, $P$ has to learn an adversarial policy $\pi_{adv}$ such that its accumulated discounted rewards $G_{t|\pi_{adv}}=\sum^{T}_{\tau = t}\gamma^{\tau-t}r(x_{\tau},a_{\tau},x_{\tau+1}, {h_t})$ is maximized.  Second, $I$ requires to learn an optimal $\theta_{I}$ such that Eq.~(\ref{eq::inv_reward}) is minimized.  Minimizing $L_I$ (i.e., $r_t$) leads to a decreased $G_{t|\pi_{adv}}$, forcing $P$ to enhance $\pi_{adv}$ to explore more difficult samples to increase $G_{t|\pi_{adv}}$.  This implies that $P$ is motivated to concentrate on {discovering} $I$'s weak points {in the state space}, instead of randomly or {even repeatedly} collecting ineffective training samples for $I$.  
Training $I$ with hard samples not only accelerates its learning progress, but also helps to boost its performance. {First of all, hard samples that provide higher prediction errors in Eq.~(\ref{eq::inv_reward}) are unfamiliar to $I$, offering a direction for $P$ to explore in the environment. % state space 
Such a directed exploration scheme enables our adversarial active exploration to discover unfamiliar data more rapidly than random exploration.  Moreover, those hard samples that require more gradient steps to learn tend to result in larger errors of $I$ than samples which can be learned in just a few steps, which in turn help to expedite the learning progress of $I$.  Data samples causing negligible errors of $I$ do not provide the same benefits, as they are already familiar to $I$.  Our method is able to discover those hard samples in an environment and focus on learning them, which is similar to what boosting algorithms~\cite{schapire1999brief} do.} 

Comparing to active learning and curiosity-driven exploration, our adversarial active exploration demonstrates several strengths on inverse dynamics model learning. First of all, active learning~\cite{settles2009active} does not have a control policy to acquire the informative data in an environment, while {our method} casts the data acquisition process for inverse dynamics models as a decision process and solves it using DRL. {Second, the curiosity-driven exploration approach~\cite{pathak2018zeroshot} collects samples not directly tailored for $I$`s learning, while our method gathers samples that directly influence $I$`s learning progress.  Moreover, the incentives for exploration in curiosity-based approaches disappear rapidly when the forward dynamics model quickly learns to perform correct predictions, which may further undermine the training progress of their inverse dynamics models. On the contrary, our method continuously explores the environment until $I$ converges. A more detailed comparison is presented in Section~\ref{subsec::exp_arm}.}

\subsection{Training Methodology}
\label{sec::adv_exp}
We describe the training methodology of our adversarial active exploration method by a pseudocode presented in Algorithm~\ref{alg::adv_exp}. Assume that $P$'s policy $\pi_{adv}$ is parameterized by a set of trainable parameters $\theta_P$, and is represented as $\pi_{adv}(a_t|x_t, \theta_P)$. We create two buffers $Z_P$ and $Z_I$ for storing the training samples of $P$ and $I$, respectively. In the beginning, $Z_P$, $Z_I$, $\mathcal{E}$, $\theta_P$, $\theta_I$, $\pi_{adv}$, as well as a timestep cumulative counter $c$ are initialized. A number of hyperparameters are set to appropriate values, including the number of iterations $N_{iter}$, the number of episodes $N_{episode}$, the horizon $T$, as well as the update period $\mathcal{T}_{P}$ of $\theta_P$.  At each timestep $t$, $P$ perceives the current state $x_t$ from $\mathcal{E}$, takes an action $a_t$ according to $\pi_{adv}(a_t|x_t, \theta_P)$, and receives the next state $x_{t+1}$ and a termination indicator $\xi$ (lines 9-11). $\xi$ is set to 1 only when $t$ equals $T$, otherwise it is set to 0. We then store $(x_t, a_t, x_{t+1}, \xi)$ and $(x_t, a_t, x_{t+1})$ in $Z_P$ and $Z_I$, respectively.  We update $\theta_P$ every $\mathcal{T}_{P}$ timesteps using the samples stored in $Z_P$, as shown in (lines 13-21). At the end of each episode, we update $\theta_I$ with samples drawn from $Z_{I}$ according to the loss function $L_I$ defined in Eqs.~(\ref{eq::inv_loss}) and~(\ref{eq::inv_reward}) (line 23).

\begin{algorithm}[t]
\scriptsize
\caption{Adversarial Active Exploration}
\label{alg::adv_exp}
\begin{algorithmic}[1]
\State Initialize $Z_{P}$, $Z_{I}$, $\mathcal{E}$, and model parameters $\theta_{P}$ \& $\theta_{I}$
\State Initialize $\pi_{adv}(a_t|x_t, \theta_P)$%RL policy $\pi(x; \theta_{RL})$
\State Initialize the timestep cumulative counter $c=0$
\State Set $N_{iter}$, $N_{episode}$, $T$, and $\mathcal{T}_{P}$
\For{iteration $i=1$ to $N_{iter}$}
	\For{episode $e=1$ to $N_{episode}$}
    	%\State $x_0 = reset(\mathcal{E})$
    	\For{timestep $t=0$ to $T$}
        	\State $P$ perceives $x_t$ from $\mathcal{E}$, and predicts an action $a_t$ according to $\pi_{adv}(a_t|x_t, \theta_P)$
            \State $x_{t+1}$ = $\mathcal{E}(x_t, a_t)$
            \State $\xi = \mathbbm{1}[t == T]$
            \State Store $(x_t, a_t, x_{t+1}, \xi)$ in $Z_{P}$
            \State Store $(x_t, a_t, x_{t+1})$ in $Z_{I}$
            \If{($c$ $\mod$ $\mathcal{T}_{P}$) == 0}
            	\State Initialize an empty batch $B$
                \State Initialize a recurrent state $h_t$
            	\For{$(x_t, a_t, x_{t+1}, \xi)$ in $Z_{P}$}
                	%\If{$d$}
                    %	\State $reset\_recurrent\_state(h)$
                    %\EndIf
                    \State Evaluate $\hat{a_t} = I(x_t, x_{t+1}|h_t, \theta_{I})$ (calculated from Eq.~(\ref{eq::a_t}))
                	\State Evaluate $r(x_t, a_t, x_{t+1}, h_t) = L_I(a_t, \hat{a_t}| \theta_{I})$ (calculated from Eq.~(\ref{eq::inv_reward}))
                    %\State $h = udate\_recurrent\_state(h)$
                	\State Store $(x_t, a_t, x_{t+1}, r_t)$ in $B$
                \EndFor
            	\State Update $\theta_{P}$ with the gradient calculated from the samples of $B$
                \State Reset $Z_{P}$
            \EndIf
           % \State $x_t = x_{t+1}$ 
            \State $c = c + 1$
        \EndFor
    \EndFor
   \State Update $\theta_{I}$ with the gradient calculated from the samples of $Z_{I}$ (according to Eq.~(\ref{eq::inv_loss}))
\EndFor
\State \textbf{end}
\end{algorithmic}
\end{algorithm}

\subsection{Stabilization Technique}
\label{sec::tech_stab}
Although adversarial active exploration is effective in collecting hard samples, it requires additional adjustments if $P$ becomes too strong such that the collected samples are too difficult for $I$ to learn.
Overly difficult samples lead to a large magnitudes in gradients derived from $L_I$, which in turn cause a performance drop in $I$ and instability in its learning process. We analyze this phenomenon in greater detail in Section~\ref{exp::abl}. To tackle the issue, we propose a training technique that reshapes $r_t$ as follows:
\begin{equation}
\footnotesize
\label{eq::stabilization}
	r_t := -|r_t - \delta|,
\end{equation}
where $\delta$ is a pre-defined threshold value. This technique poses a restriction on the range of $r_t$, driving $P$ to gather moderate samples instead of overly hard ones. Note that the value of $\delta$ affects the learning speed and the final performance. We plot the impact of $\delta$ on the learning curve of $I$ in Section~\ref{exp::abl}. We further provide an example in {our supplementary} material to visualize the effect of this technique.

\section{Experimental Results}
\label{sec::exp}
%\TODO{Don't say ours is better than "curiosity-driven". Say "forward error driven"}
In this section, we present experimental results for a series of robotic control tasks, and validate that (i) our method is effective for common robotic arm control and in-hand manipulation tasks; (ii) our method is effective in environments with high-dimensional action spaces; (iii) our method is more data efficient than the baseline methods.  We first introduce our experimental setup. Then, we report our experimental results on a number of robotic arm and hand manipulation tasks. Finally, we present a comprehensive set of ablative analysis to validate the effectiveness for each of our design decisions.

% we conduct a series of experiments on robotic tasks to verify the following arguments, which compare our method against autonomous data collection strategies proposed in previous relevant works. \textcolor{red}{outline}
%In this section, we conduct a series of experiments on robotic tasks to support the following arguments. The following argument compare our method and the autonomous data-collection strategies proposed in the previous works.
% \begin{enumerate}[label=(\roman*)]
% \item Our method is more effective in robotic arm manipulation tasks, even when working with visual observations.
% \item Our method is particularly effective in the environment with high-dimensional action spaces.
% \item Our method is more data-efficient.
% \item Our method is more robust against action space noise.
% \end{enumerate}
% In the subsequent sections, we first introduce our experimental setup. Next, we analyze the experimental results on several robotic tasks. Finally, we present an ablative analysis to justify each of our design choices.

\subsection{Experimental Setup}
%\TODO{Elaborate the experimental settings}
We first describe the environments and tasks. Next, we explain the evaluation procedure and the method for collecting expert demonstrations. We then walk through the baseline approaches used in our comparisons.  {Please note that for all of our experiments, we train our DRL agent using PPO~\cite{schulman2017proximal}.}

%We begin with explaining the environments and tasks. We then explain the evaluation procedure and the collection of expert demonstrations. Finally, we introduce the baselines for comparison. 

% \textcolor{red}{(Is 'imitator' a appropriate term?)}
\subsubsection{Environments and Tasks}  
\label{sec::setup_env_task}
We evaluate our method on a number of robotic arm and hand manipulation tasks via OpenAI gym~\citep{Brockman2016OpenAIGym} environments simulated by the MuJoCo~\citep{Todorov2012mujoco} physics engine. We use the Fetch and Shadow Dexterous Hand~\citep{plappert2018multi} for the arm and hand manipulation tasks, respectively. For the arm manipulation tasks, which include \textit{FetchReach}, \textit{FetchPush}, \textit{FetchPickAndPlace}, and \textit{FetchSlide}, $I$ takes as inputs the positions and velocities of a gripper and a target object. It then infers the gripper's action in 3-dimensional space to manipulate it. For the hand manipulation task \textit{HandReach}, the inverse dynamics model takes as inputs the positions and velocities of the fingers of a robotic hand, and determines the velocities of the joints to achieve the goal. The detailed description of the above tasks is specified in~\cite{plappert2018multi}. For the detailed configurations of these tasks, please refer to our {supplementary} material.

\subsubsection{Evaluation Procedure}
\label{sec::eval_proc}
The primary objective of our experiments is to demonstrate the efficiency of the proposed adversarial active exploration in collecting training data (in a self-supervised manner) for inverse dynamics models.  We compare our method against a number of data collection methods (referred to as "baselines" or "baseline methods") described in Section~\ref{subsubsec::baseline_methods}.  As different baseline methods employ different data collection strategies, the learning curves of {their inverse dynamics models also vary for different cases.}  For a fair comparison, the model architecture of the inverse dynamics model and the amount of training data are fixed for all cases.  All of the experimental results are evaluated and averaged over 20 trials, corresponding to 20 different random initial seeds.  In each trial, we train an inverse dynamics model by the training data collected by a single data collection method. We periodically evaluate the inverse dynamics model when every 10K training samples are collected. At the beginning of each episode of evaluation, the inverse dynamics model receives a sequence of states $\{{\hat{x}_{1}, \hat{x}_{2},\cdots,\hat{x}_{T}}\}$ from a successful expert demonstration. At each timestep $t$, the inverse dynamics model infers an action $\hat{a}_t$ from an expert state $\hat{x}_{t+1}$ and its current state $x_t$ by Eq.~(\ref{eq::a_t}). The evaluation is performed by averaging the success rates of reaching $\hat{x}_{T}$ over 500 episodes. The configuration of the inverse dynamics model and the hyperparameters are summarized in the {supplementary} material.

\subsubsection{Expert Demonstrations for Evaluation} 
For each task mentioned in Section~\ref{sec::setup_env_task}, we first randomly configure task-relevant settings (e.g., goal position, initial state, etc.). \cyl{The configuration details of these tasks are specified in the codebase of OpenAI gym\footnote{https://github.com/openai/gym}.} We then collect demonstrations from non-trivial and successful episodes performed by a pre-trained expert agent~\citep{andrychowicz2017hindsight}. Please note that the collected demonstrations only contain sequences of states. 
%The interested reader is referred to our {supplementary} material for 
The implementation details of the expert agent and the methodology for filtering out trivial episodes are presented and discussed in detail in our {supplementary} material.
%For each task mentioned at Section~\ref{sec::setup_env_task}, we randomly configure the task-relevant settings (i.e. destination, initial state), collecting those non-trivial and successful episodes generated by a pre-trained expert agent~\cite{andrychowicz2017hindsight}. The implementation detail of the pre-trained expert agent is leaved in Supplementary material. To avoid the imitator succeeds on the task without taking actions, we filtered out the trivial episodes according to the task-specific schemes, which are described in Supplementary materials. Note that the demonstrations only include experts’ states, excluding the expert’s actions. 

\subsubsection{Baseline Data Collection Methods}
\label{subsubsec::baseline_methods}
We compare our proposed methodology with the following four baseline methods in our experiments.
\begin{itemize}[itemsep=.25ex]
\item \textit{Random}: This method collects training samples by random action sampling. We consider it to be an important baseline method because of its simplicity and prevalence in a number of research works on inverse dynamics models learning~\citep{pathak2018zeroshot,nair2017combining,agrawal2016learning}.
%This strategy collects the training data by taking a action randomly at each timestep. We consider it as a baseline since it is the simplest data-collection strategy which is broadly used in several works~\cite{agrawal2016learning,nair2017combining,pathak2018zeroshot}.
%\textit{Random} to pursue the given demonstrations. Then it collects the trajectories generated by the imitator, training the imitator with these data finally. Implementation detail can be found in Supplementary material.
\item \textit{Demo}: This method trains the inverse dynamics model directly with expert demonstrations. {Since expert demonstrations comprise only successful trajectories}, this method serves as a baseline with human priors similar to \cite{nair2017combining}.

% It serves as the performance upper bound, as the training data is the same as the testing data for this method.
%This method trains the inverse dynamic model with the expert’s demonstration directly. It provide us the performance upper bound of an architecture since we consider it is the optimal data-collection strategy as the training data are the same as the testing data.
\item \textit{Curiosity}: This method incentivizes a DRL agent to collect samples that lead to large errors of its forward dynamics model~\citep{pathak2018zeroshot,pathak2017curiosity}. Unlike the original implementation, we replace its DRL algorithm with PPO. This is also an important baseline due to its effectiveness in~\cite{pathak2018zeroshot}.
%This method train a RL agent with only intrinsic rewards, collecting the training data for inverse dynamic model with this RL agent.  The intrinsic rewards are generated from the losses of the forward dynamic model jointly trained with the RL agent. Note that differing from the original implementation in~\cite{pathak2017curiosity}, we replace the RL algorithm used in~\cite{pathak2017curiosity} with PPO as the training should be limited to single thread for fair comparison with the other methods. We take it as one of baselines for comparison as it demonstrates the effectiveness in~\cite{pathak2018zeroshot}.
\item \textit{Noise}~\citep{plappert2018parameter}:  In this method, noise is injected to the parameter space of a DRL agent to encourage exploration~\citep{plappert2018parameter}. {We include this baseline as a reference to validate if a standard DRL exploration strategy can provide positive impact on inverse dynamics model learning.}
%The detailed implementation can be found in our \textcolor{red}{supplementary} material. 
% We include this method due to its superior performance and data efficiency in many DRL tasks.
%, including MuJoCo. %It is in our opinion worth investigating whether an exploration strategy for RL can improve the data-efficiency in self-supervised IL.
%The training samples of the imitator are collected by a RL agent trained with the method proposed in~\cite{plappert2018parameter}. In this method, noise is injected to the parameter space of the RL agent to encourage the RL agent explore the environment. Note that the exploratory behavior fully depends on parameter space nose and the RL agent receives no rewards. The implementation detail is described in Supplementary material. We take this method for comparison since it demonstrates the superior performance and data-efficiency on RL tasks in several domains including MuJoCo. It is worth investigating whether the exploration strategy for RL can improve the data-efficiency in self-supervised imitation learning. 
\end{itemize} 

\subsection{Performance Comparison in Robotic Arm Manipulation Tasks}
\label{subsec::exp_arm}
\begin{figure}[t]
\centering
\vspace{-3ex}
\includegraphics[width=\linewidth]{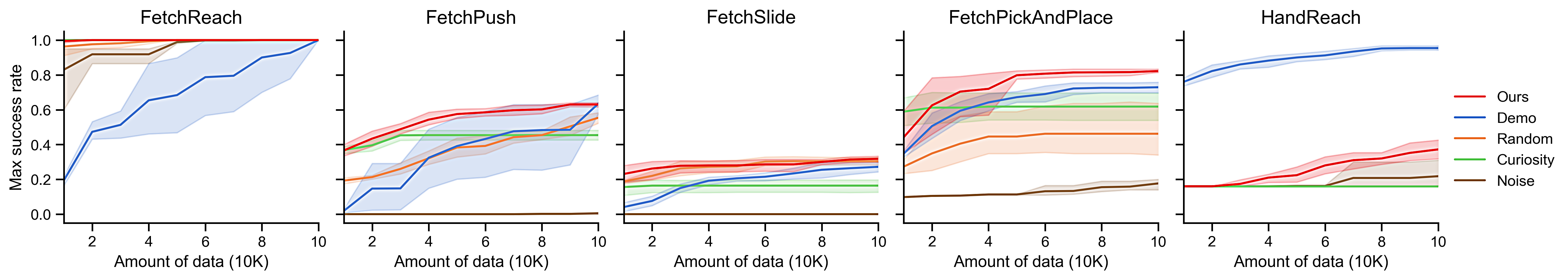}
\caption{Performance comparison of robotic arm and hand tasks.}
\label{figure::perf_robot_arm_low}
\end{figure}

%We benchmark our method on multiple robotic arm manipulation tasks: \textit{FetchReach}, \textit{FetchPush}, \textit{FetchPickAndPlace}, and \textit{FetchSlide}. 
We compare the performance of the proposed method and the baselines on the robotic arm manipulation tasks described in Section~\ref{sec::setup_env_task}.
%As opposed to discrete control domains, these tasks are especially challenging, as the sample complexity grows in continuous control domains.
%Furthermore, the imitator may not have the complete picture of the environment dynamics, increasing its difficulty to learn an inverse dynamics model. In \textit{FetchSlide}, for instance, the movement of the object on the slippery surface is affected by both friction and the force exerted by the gripper. It thus motivates us to investigate whether the proposed method can help overcome the challenge. 
We discuss the experimental results and plot them in Fig.~\ref{figure::perf_robot_arm_low}.  All of the results are obtained by following the procedure described in Section~\ref{sec::eval_proc}.  The shaded regions in Fig.~\ref{figure::perf_robot_arm_low} represent the confidence intervals.
Fig.~\ref{figure::perf_robot_arm_low} plots the learning curves for all of the methods.  In all of the tasks, our method yields superior or comparable performance to the baselines except for \textit{Demo}, which is trained directly with expert demonstrations (i.e. human priors). In \textit{FetchReach}, it can be seen that every method achieves a success rate of 1.0. This implies that it does not require a sophisticated exploration strategy to learn an inverse dynamics model in an environment where the dynamics is relatively simple. It should be noted that although all methods reach the same final success rate, ours learns significantly faster than \textit{Demo}. In contrast, in \textit{FetchPush}, our method is comparable to \textit{Demo}, and demonstrates superior performance to the other baselines. Our method also learns drastically faster than all the other baselines, which confirms that the proposed strategy does improve the performance and efficiency of inverse dynamics model learning. Our method is particularly effective in tasks that require complex control. In \textit{FetchPickAndPlace}, for example, our method surpasses all the other baselines. However, all methods including \textit{Demo} fail to learn a successful inverse dynamics model in \textit{FetchSlide}, which suggests that it is difficult to train an inverse dynamics model when the outcome of an action {is partially affected by the unobservable factors of an environment}. In \textit{FetchSlide}, the movement of the object on the slippery surface is affected by both friction and the force exerted by the gripper. It is worth noting that \textit{Curiosity} loses to \textit{Random} in \textit{FetchPush} and \textit{FetchSlide}, and \textit{Noise} performs even worse than these two methods in all of the tasks. We therefore conclude that \textit{Curiosity} and the parameter space noise strategy cannot be directly applied to inverse dynamics model learning. {Moreover, Fig.~\ref{figure::perf_robot_arm_low} shows that the performance of \textit{Curiosity} saturates rapidly, indicating that \textit{Curiosity} is unable to continuously explore the environment until $I$ converges}. In addition to the quantitative results presented above, we further discuss the empirical results qualitatively. Please refer our supplementary material for a detailed description of the results.

% \paragraph{High-dimensional state spaces.} Fig.~\ref{figure::perf_robot_arm_high} plots the learning curves of all methods in high-dimensional state spaces. It can be seen that our method performs significantly better than the other baseline methods in most of the tasks, and is comparable to \textit{Demo}. In \textit{FetchPickAndPlace}, our method is the only one that learns a successful inverse dynamics model. Similar to the results in Fig.~\ref{figure::perf_robot_arm_low}, \textit{Curiosity} is no better than \textit{Random} in high-dimensional state spaces. Please note that we do not include \textit{Noise} in Fig.~\ref{figure::perf_robot_arm_high}  as it performs worse enough already in low-dimensional settings. %Therefore we believe that it will fail in visual imitation tasks as well.

% \begin{figure}[t]
% \centering
% \includegraphics[width=0.8\linewidth]{figures/fig_learn_curve_rgb_robot.png}
% \caption{Performance comparison of robotic arm tasks in high-dimensional state  spaces.}
% \label{figure::perf_robot_arm_high}
% \end{figure}

% Overall, our method demonstrates superior performance in terms of both efficiency and success rates. We are convinced that our method is more effective in and suitable for complex robotic tasks than other self-supervised IL methods.

\subsection{Performance Comparison in Robotic Hand Manipulation Task}
\label{subsec::exp_hand}
%Apart from robotic arm tasks, we investigate into a more complicated topic: controlling a robot hand. In \textit{HandReach}, high-dimensional action spaces make learning an inverse dynamics model more challenging, which motivates us to evaluate the effectiveness of our method in such a setting. 
Fig.~\ref{figure::perf_robot_arm_low} plots the learning curves for each of the methods considered. 
%All the experimental results are obtained by following the procedure in Section~\ref{sec::eval_proc}. 
Please note that \textit{Curiosity}, \textit{Noise} and our method are pre-trained with 30K samples collected by random exploration, as we observe that these methods on their own suffer from large errors in an early stage during training, which prevents them from learning at all. After the first 30K samples, they are trained with data collected by their exploration strategy instead. From the results in Fig.~\ref{figure::perf_robot_arm_low}, it can be seen that \textit{Demo} easily stands out from the other methods as the best-performing model, surpassing them all by a considerable extent. Although our method is not as impressive as \textit{Demo}, it significantly outperforms all of the other baseline methods, achieving a success rate of 0.4 while the others are still stuck at around 0.2. 

The reason that the inverse dynamics models trained by the autonomous data-collection strategies discussed in this paper (including ours and the other baselines) are not comparable to the \textit{Demo} baseline in the \textit{HandReach} task is primarily due to the high-dimensional action space.  It is observed that the data collected by the autonomous data-collection strategies only cover a very limited range of the state space in the \textit{HandReach} environment.  Therefore, the inverse dynamics models trained with these data only learn to predict trivial poses, leading to the poor success rates presented in Fig.~\ref{figure::perf_robot_arm_low}.

\subsection{Ablative Analysis}
\label{exp::abl}
In this section, we provide a set of ablative analysis.  We examine the effectiveness of our method by an investigation of the training loss distribution, the stabilization technique, and the influence of $\delta$.  Please note that the value of $\delta$ is set to 1.5 by default, as described in our supplementary material.
%First of all, we present the distribution of training errors of collected samples. Then we show the effectiveness of the stabilization technique. Finally, we discuss the influence of various $\delta$.
    
\begin{figure}[t!]
\centering
%\vspace{-3ex}
\includegraphics[width=\linewidth]{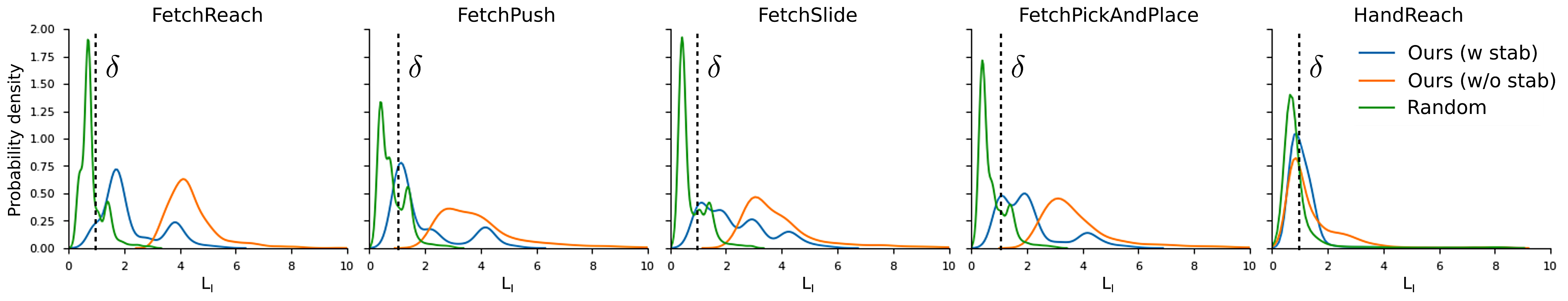}
\caption{PDFs of $L_I$  for the first 2K training batches.}
\label{figure::error_stab}
\end{figure}

\paragraph{Training loss distribution.} Fig.~\ref{figure::error_stab} plots the probability density function (PDF) of $L_I$ (derived from Eq.~(\ref{eq::inv_loss})) by kernel density estimation (KDE) for the first 2K training batches during the training phase.  The vertical axis corresponds to the probability density, while the horizontal axis represents the scale of $L_I$.  The curves \textit{Ours~(w stab)} and \textit{Ours~(w/o stab)} represent the cases where the stabilization technique described in Section~\ref{sec::tech_stab} is employed or not, respectively.  We additionally plot the curve \textit{Random} in Fig.~\ref{figure::error_stab} to highlight the effectiveness of our method.  It can be observed that both \textit{Ours~(w stab)} and \textit{Ours~(w/o stab)} concentrate on notably higher loss values than \textit{Random}.  This observation implies that adversarial active exploration does explore hard samples for inverse dynamics model.

%our method with and without the use of the stabilization technique. The vertical axis corresponds to the number of samples, while the horizontal axis corresponds to $L_I$. The curves in Fig.~\ref{figure::error_stab} are smoothed by kernel density estimation. It can be seen that both \textit{Ours(w stab)} and \textit{Ours(w/o stab)} concentrate on notably higher values than \textit{Random}. This indicates that the adversarial exploration strategy does help collect hard samples for the inverse dynamics model.

\begin{figure}[t]
\centering
\includegraphics[width=\linewidth]{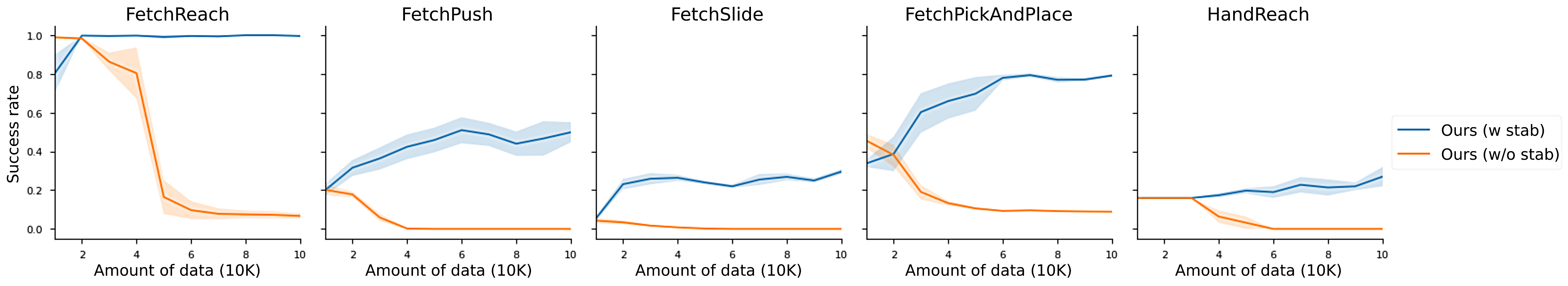}
\caption{
Learning curves w/ and w/o the stabilization technique.
%Performance comparison of w/ and w/o stabilization technique in low-dimensional observation spaces.
}
\label{figure::perf_stab_low}
\end{figure}

%the errors of collected samples of \textit{Ours (w/o stab)} concentrates in the extremely higher error part than the others. It indicates that our method does collect the hard samples for the inverse dynamic model.
\paragraph{Validation of the stabilization technique.}  We validate the proposed stabilization technique in terms of the PDF of $L_I$ and the learning curve of the inverse dynamics model, and plot the results in Figs.~\ref{figure::error_stab} and~\ref{figure::perf_stab_low}, respectively. From Fig.~\ref{figure::error_stab}, it can be observed that the modes of \textit{Ours~(w stab)} are lower than those of \textit{Ours~(w/o stab)} in most cases, implying that the stabilization technique indeed motivates the DRL agents to favor those moderately hard samples.  We also observe that for each of the five cases, the mode of \textit{Ours~(w stab)} is close to the value of $\delta$ (plotted in a dotted line), indicating that our reward structure presented in Eq.~(\ref{eq::stabilization}) does help to regulate $L_I$ (and thus $r_t$) to be around $\delta$.  To further demonstrate the effectiveness of the stabilization technique, we compare the learning curves of \textit{Ours~(w stab)} and \textit{Ours~(w/o stab)} in Fig.~\ref{figure::perf_stab_low}.  It is observed that for the initial 10K samples of the five cases, the success rates of \textit{Ours~(w/o stab)} are comparable to those of \textit{Ours~(w stab)}.  However, their performance degrade drastically during the rest of the training phase.  This observation confirms that the stabilization technique does contribute significantly to our adversarial active exploration.

Although most of the DRL works suggest that the rewards should be re-scaled or clipped within a range (e.g., from -1 to 1), the unbounded rewards do not introduce any issues during the training process of our experiments.  The empirical rationale is that the rewards received by the DRL agent are regulated by Eq.~(\ref{eq::stabilization}) to be around $\delta$, as described in Section~\ref{exp::abl} and depicted in Fig.~\ref{figure::error_stab}.  Without the stabilization technique, however, the learning curves of the inverse dynamics model degrade drastically (as illustrated in Fig.~\ref{figure::perf_stab_low}). %even if the reward clipping technique is applied. 

%they suffer from a significant degradation in performance for the rest of the training phase. 

%\textit{Ours~(w stab)} has a lower mean loss than \textit{Ours~(w/o stab)}, which implies that the stabilization technique successfully guides the DRL agent to favor those moderately hard samples. We also observe that the center of loss distribution for \textit{Ours~(w stab)} is close to the value of $\delta$, as shown in Fig.~\ref{figure::error_stab}, confirming that our reward structure guides data collection by $\delta$. To further demonstrate the effectiveness of the stabilization technique, we plot the learning curves of \textit{Ours~(w stab)} and \textit{Ours~(w/o stab)} in Fig.~\ref{figure::perf_stab_low}. Although \textit{Ours~(w/o stab)} is comparable to \textit{Ours~(w stab)} for the initial 10K samples, it suffers from a significant degradation in performance for the rest of the training phase. This result indicates that the stabilization technique does improve the overall performance of our method.

\begin{figure}[t!]
\centering
\includegraphics[width=\linewidth]{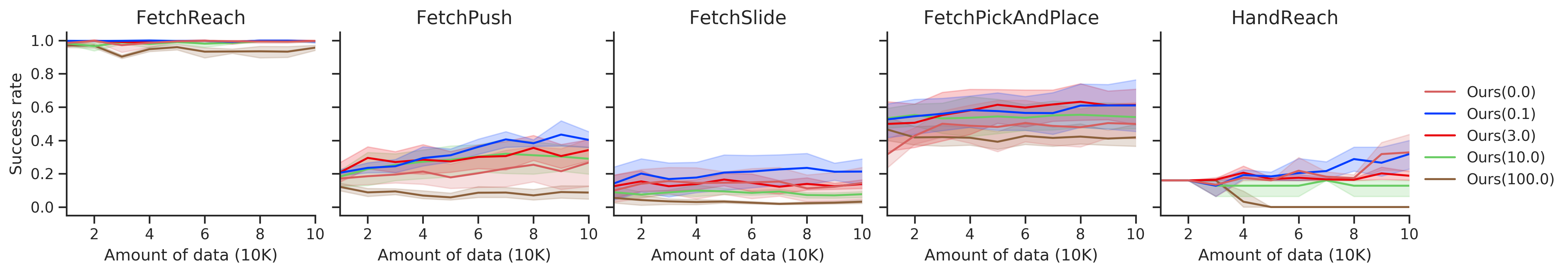}
\caption{Performance comparison for different values of $\delta$.}
\label{figure::perf_stab_delta}
\end{figure}

%From Fig. XX, it can be seen that the errors of collected samples of \textit{w stab} concentrates in a moderate hard part, which indicates that stabilization technique successfully guide the data-collection toward the moderate hard data. Moreover, the center of the distribution of \textit{w stab} is close to the selected $\delta$. It shows that the our reward structure can guide the data-collection by $\delta$ setting. To justify stabilization leads to better performance, we plot the learning curves of different settings in Fig. XX. The meanings of \textit{w stab} and \textit{w/o stab} in Fig.XX  are the same as Fig. XX. It can be observed that \textit{w/o stab} obtains a comparable performance with initial 10K data, however, degrades in the consequent of training progress. This result indicates that the stabilization technique indeed improves the performance 
\paragraph{Influence of $\delta$.} 
Fig.~\ref{figure::perf_stab_delta} compares the learning curves of the inverse dynamics model for different values of $\delta$.  For instance, \textit{Ours(0.1)} corresponds to $\delta = 0.1$. %In the legend, the value within the parentheses indicates the value of $\delta$ (e.g. Ours(0.1) corresponds to $\delta = 0.1$). 
It is observed that for most of the tasks, the success rates drop when $\delta$ is set to an overly high or low value (e.g., 100.0 or 0.0), suggesting that a moderate value of $\delta$ is necessary for the stabilization technique.  The value of $\delta$ can be adjusted dynamically by the adaptive scaling technique presented in  \cite{plappert2018parameter}, which is left as our future direction. 

% From the above analysis, we conclude that our adversarial active exploration  is effective in collecting difficult training data for the inverse dynamics model.  The analysis also validates that our stabilization technique indeed leads to superior performance, and is capable of guiding the DRL agent to collect moderately hard samples. This enables the inverse dynamics model to pursue a stable learning curve.

%drops with a overly high or low $\delta$. This result suggests that the proper setting of $\delta$ is crucial. However, it could be dynamically adjusted by the adaptive scaling technique adopted in \citet{plappert2018parameter} in the future work.

%Fig.~\ref{figure::perf_stab_delta} plots the learning curves of our methods using $\delta = 0.1$ and $\delta = 3.0$. From the experimental results, we observe that \textit{Ours(0.1)} and \textit{Ours(3.0)} perform comparably, which means that the choice of $\delta$ has little influence on the model's performance.

\section{Conclusion}
\label{sec::conclusion}
In this paper, we presented an adversarial active exploration, which consists of a DRL agent and an inverse dynamics model competing with each other for efficient data collection.  The former is encouraged to actively collect difficult training data for the latter, such that the training efficiency of the latter is significantly enhanced.  Experimental results demonstrated that our method substantially improved the data collection efficiency in multiple robotic arm and hand manipulation tasks, and boosted the performance of the inverse dynamics models. 
% Camera ready
The comparisons with the forward dynamics errors driven approach~\cite{pathak2018zeroshot} validated that our method is more effective for training inverse dynamics models.  We further showed that our method is generalizable to environments with high-dimensional action spaces. Finally, we provided a set of ablative analysis to justify each of our design decisions.
% Original
% The comparisons with curiosity-driven approach further validated that our method is more effective for training inverse dynamics models. In addition, we showed that our method is generalizable to environments with high-dimensional action spaces. Finally, we provided a set of ablative analysis to further justify each of our design decisions.

\section*{Acknowledgment}
This work was supported by the Ministry of Science and Technology (MOST) in Taiwan under grant nos. MOST 108-2636-E-007-010 (Young Scholar Fellowship Program) and MOST 108-2634-F-007-002.  Z.-W. Hong and C.-Y. Lee acknowledge the financial support from MediaTek Inc., Taiwan.  The authors would also like to acknowledge the donation of the Titan XP GPUs from NVIDIA Corporation and the Titan V GPUs from NVIDIA AI Technology Center used in this research work.

% no \bibliographystyle is required, since the corl style is automatically used.
\bibliography{example}  % .bib

\appendix
\begin{appendices}

% Change section numbering
%\renewcommand{\thesection}{S.\arabic{section}}
\section{Qualitative Analysis of Robotic Arm Manipulation Tasks}
In addition to the quantitative results presented above, we further discuss the empirical results qualitatively. Through visualizing the training progress, we observe that our method initially acts like \textit{Random}, but later focuses on interacting with the object in \textit{FetchPush}, \textit{FetchSlide}, and \textit{FetchPickAndPlace}. This phenomenon indicates that adversarial active exploration  naturally gives rise to a curriculum that improves the learning efficiency, which resembles curriculum learning~\citep{bengio2009curriculum}. Another benefit that comes with the phenomenon is that data collection is biased towards interactions with the object. Therefore, the DRL agent concentrates on collecting interesting samples that has greater significance, rather than trivial ones. For instance, the agent prefers pushing the object to swinging the robotic arm. 
%It also suggests that the adversarial exploration strategy approximates the human-biased sampling strategy in~\citet{agrawal2016learning,nair2017combining}, but does not require any form of human supervision. 
On the other hand, although \textit{Curiosity} explores the environment very thoroughly in the beginning by stretching the arm into numerous different poses, it quickly overfits to one specific pose. This causes its forward dynamics model to keep maintaining a low error, making it less curious about the surroundings. 
%Based on this finding, we reason that following the convergence of the forward dynamics model, the DRL agent becomes less curious about its surroundings, thus making it linger in certain areas of the environment. 
Finally, we observe that the exploratory behavior of \textit{Noise} does not change as frequently as ours, \textit{Random}, and \textit{Curiosity}. We believe that the method's success in the original paper~\citep{plappert2018parameter} is largely due to extrinsic rewards. In the absence of extrinsic rewards, however, the method becomes less effective and unsuitable for data collection, especially in inverse dynamics model learning.

\section{Proximal Policy Optimization (PPO)}
\label{appendix:ppo}
We employ PPO \citep{schulman2017proximal} as the RL agent responsible for collecting training samples because of its ease of use and good performance.
PPO computes an update at every timestep that minimizes the cost function while ensuring the deviation from the previous policy is relatively small. One of the two main variants of PPO is a clipped surrogate objective expressed as:
\begin{equation*}
L^{CLIP}(\theta) = \mathbb{E} \left[ \frac{\pi_{\theta}(a|s)}{\pi_{\theta_{old}}(a|s)}\hat{A}, \text{clip}(\frac{\pi_{\theta}(a|s)}{\pi_{\theta_{old}}(a|s)}, 1-\epsilon, 1+\epsilon)\hat{A}) \right],
\end{equation*}
where $\hat{A}$ is the advantage estimate, and $\epsilon$ a hyperparameter. The clipped probability ratio is used to prevent large changes to the policy between updates. The other variant employs an adaptive penalty on KL divergence, given by:
\begin{equation*}
L^{KLPEN}(\theta) = \mathbb{E} \left[ \frac{\pi_{\theta}(a|s)}{\pi_{\theta_{old}}(a|s)}\hat{A} - \beta KL\left[\pi_{\theta_{old}}(\cdot|s), \pi_{\theta}(\cdot|s)\right]\right],
\end{equation*}
where $\beta$ is an adaptive coefficient adjusted according to the observed change in the KL divergence. In this work, we employ the former objective due to its better empirical performance.

\section{Implementation Details of Inverse Dynamics Model}
In the experiments, the inverse dynamics model $I(x_t, x_{t+1}|h_t, \theta_{I})$ of all methods employs the same network architecture. We use 3 Fully-Connected (FC) layers with 256 hidden units followed by $tanh$ activation units. We then feed the extracted features from the FC layers to Long-Short Term Memory (LSTM)~\cite{Hochreiter:1997:LSM:1246443.1246450}.
% In the experiments, the inverse dynamics model $I(x_t, x_{t+1}|h_t, \theta_{I})$ of all methods employs the same network architecture, which is illustrated in Fig. \ref{figure::inv_model_archi}. "FC-256" and "FC-$|\mathcal{A}|$" indicate a fully connected layer with 256 hidden units and with $|\mathcal{A}|$ hidden units respectively. "LSTM-256" means a Long-Short Term Memory (LSTM) \citep{gers1999learning} layer with 256 hidden units. "tanh" indicates a hyperbolic activation unit, and "Concat" means concatenation operation.  The hyperparameters for training the inverse dynamics model can be found in Table. \ref{table::hyperparam}.
%All methods employ the same network architecture for the inverse dynamics model in the experiments. The network architecture of the inverse dynamic model $I(x_t, x_{t+1}|h_t, \theta_{I})$ are illustrated in Fig. \ref{figure::inv_model_archi}. "FC-256" and "FC-$|\mathcal{A}|$" indicate a fully connected layer with 256 hidden units and with $|\mathcal{A}|$ hidden units respectively. "LSTM-256" means a Long-Short Term Memory (LSTM) \citep{gers1999learning} layer with 256 hidden units. "tanh" indicates hyperbolic activation unit. "Concat" means concatenation operation.\ref{figure::inv_model_archi}.  The hyperparameters for training inverse dynamic model can be found in Table. \ref{table::hyperparam}.

\section{Implementation Details of Adversarial Active Exploration}
We use the architecture proposed in \citet{schulman2017proximal}. During training, we periodically update the DRL agent with a batch of transitions as described in Algorithm. 1. We split the batch into several mini-batches, and update the RL agent with these mini-batches iteratively. The best hyperparameters are listed in Table.~\ref{table::hyperparam}~(our method). 

\section{Implementation details of \textit{Curiosity}}
Our baseline \textit{Curiosity} is implemented based on the work \citep{pathak2018zeroshot}. The authors in~\citet{pathak2018zeroshot} propose to employ a curiosity-driven RL agent~\citep{pathak2017curiosity} to improve the efficiency of data collection. The curiosity-driven RL agent takes curiosity as intrinsic reward signal, where curiosity is formulated as the error in an agent’s ability to predict the consequence of its own actions. This can be defined as a forward dynamics model:
\begin{equation}
\footnotesize
\hat{\phi}(x^\prime) = f(\phi(x), a; \theta_F),
\end{equation}
where $\hat{\phi}(x^\prime)$ is the predicted feature encoding at the next timestep, $\phi(x)$ the feature vector at the current timestep, $a$ the action executed at the current timestep, and $\theta_F$ the parameters of the forward model $f$. The network parameters $\theta_F$ is optimized by minimizing the loss function $L_F$:
\begin{equation}
\footnotesize
L_F\left(\phi(x), \hat{\phi}(x^\prime)\right) = \frac{1}{2} ||\hat{\phi}(x^\prime) - \phi(x_{t+1})||^2_2.
\end{equation}
We use the architecture proposed in \citet{schulman2017proximal}. The implementation of $\phi$ depends on the model architecture of the RL agent. For low-dimensional observation setting, we implement $\phi$ with the architecture of low-dimensional observation PPO. Note that $\phi$ does not share parameters with the RL agent in this case. The best hyperparameters settings can be found in Table. \ref{table::hyperparam}(Curiosity).

\section{Implementation Details of \textit{Noise}}
We directly apply the same architecture in \citet{plappert2018parameter} without any modification. Please refer to \citet{plappert2018parameter} for more detail.

\begin{table}[t]
\centering
\footnotesize
\begin{tabular}{@{}lc@{}} \toprule \toprule
\multicolumn{1}{c}{Hyperparameter}                                      & Value   \\ \toprule \toprule
\multicolumn{2}{c}{\textbf{Common}}                                               \\ \toprule
Batch size for inverse dynamic model update                                      & 64      \\ \hline
Learning rate of inverse dynamic model                                         & 1e-3    \\ \hline
Timestep per episode                                                    & 50      \\ \toprule
Optimizer for inverse dynamic model                                     & Adam      \\ \toprule
\multicolumn{2}{c}{\textbf{Our method}}                                           \\ \toprule
Number of batch for update inverse dynamic model                                   & 25      \\ \hline
Batch size for RL agent                                                 & 2050    \\ \hline
Mini-batch size for RL agent                                            & 50      \\ \hline
Number of training iteration ($N_{iter}$)                  				& 200     \\ \hline
Number of training episode per iteration ($N_{episode}$)  				& 10      \\ \hline
Horizon ($T$) of RL agent                                               & 50      \\ \hline
Update period of  RL agent                                              & 2050    \\ \hline
Learning rate of RL agent                                               & 1e-3    \\ \hline
Optimizer for RL agent                                                  & Adam      \\ \hline
$\delta$ of stabilization                                               & 1.5     \\ \toprule
\multicolumn{2}{c}{\textbf{Curiosity}}                                            \\ \toprule
Number of batch for update inverse dynamic model                        & 500     \\ \hline
Batch size for RL agent                                                 & 2050    \\ \hline
Mini-batch size for RL agent                                            & 50      \\ \hline
Number of training iteration ($N_{iter}$)                 				& 10      \\ \hline
Number of training episode per iteration ($N_{episode}$)  				& 200     \\ \hline
Horizon ($T$) of RL agent                                               & 50      \\ \hline
Update period of  RL agent                                              & 2050    \\ \hline
Learning rate of RL agent                                               & 1e-3    \\ \hline
Optimizer for RL agent                                                  & Adam      \\ \toprule
\multicolumn{2}{c}{\textbf{Noise}}                                                \\ \toprule
Number of batch for update inverse dynamic model                                  & 500     \\ \hline
The other hyperparameters                                              & Same as \citet{plappert2018parameter} \\ \hline \hline
\end{tabular}
\vspace{1ex}
\caption{Hyperparameters settings.}
\label{table::hyperparam}
\end{table}

\section{Implementation Details of \textit{Demo}}
We collect 1000 episodes of expert demonstrations using the procedure defined in Sec.~S9 for training \textit{Demo}. Each episodes lasts 50 timesteps. The demonstration data is in the form of a 3-tuple $(x_t, a, x_{t+1})$, where $x_t$ is the current observation, $a_t$ the action, and $x_{t+1}$ the next observation. The pseudocode for training \textit{Demo} is shown in Algorithm. S1 below. In each training iteration, we randomly sample 200 episodes, namely 10k transitions (line 4). The sampled data is then used to update the inverse dynamics model (line 5). 
\begin{algorithm*}[h]
\footnotesize
\caption{\textit{Demo}}\label{alg::demo}
\begin{algorithmic}[1]
\State Initialize $Z_{Demo}$, $\theta_{I}$
\State Set constants $N_{iter}$

\For{iter $i=1$ to $N_{iter}$}
	\State Sample 200 episodes of demonstration from $Z_{Demo}$ as $B$
   	\State Update $\theta_{I}$ with the gradient calculated from the samples in $B$ (according to Eq. 6)
\EndFor
\State \textbf{end}
\end{algorithmic}
\end{algorithm*}

\section{Configuration of Environments}
We briefly explain each configuration of the environment below. For detailed description, please refer to~\citet{plappert2018multi}.
\begin{itemize}
\item \textit{FetchReach:} Control the gripper to reach a goal position in 3D space. 
\item \textit{FetchPush:} Control the Fetch robot to push the object to a target position.
%Push the object to the target position on the rough surface by controlling the arm. The imitator cannot fully control the dynamic of the environment as the movement of the gripper may not affect the object. 
\item \textit{FetchPickAndPlace:} Control the gripper to grasp and lift the object to a goal position. This task requires a more accurate inverse dynamics model.
%Pick the object, place to the target position. In addition that the imitator cannot fully control the dynamic of the environment, picking and placing require a more accurate inverse dynamic model. 
\item \textit{FetchSlide:} Control the robot to slide the object to a goal position. The task requires an even more accurate inverse dynamics model, as the object's movement on the slippery surface is hard to predict. %Slide the object to the target position on the slippery surface by controlling the robotic arm. The environment dynamic is not fully controlled by the imitator and sliding behavior requires an accurate inverse dynamic model. Most importantly, the movement of the object on the slippery surface is hard to predict, which makes training an inverse dynamic model more difficult. 
\item \textit{HandReach:} Control the Shadow Dextrous Hand to reach a goal hand pose. The task is especially challenging due to high-dimensional action spaces. %Manipulate the fingers into the specific pose. Though the environment dynamic is fully controlled by the imitator, it is a challenging task for the imitator due to high-dimensional action space. 
\end{itemize}
\section{Setup of Expert Demonstration}
We employ Deep Deterministic Policy Gradient combined with Hindsight Experience Replay (DDPG-HER) \citep{andrychowicz2017hindsight} as the expert agent. For training and evaluation, we run the expert to collect transitions for 1000 and 500 episodes, respectively. To prevent the inverse dynamics model from succeeding in the task without taking any action, we only collect successful and non-trivial episodes generated by the expert agent. Non-trivial episodes are filtered out based on the following task-specific schemes:
%We take the expert's agent as Hindsight Experience Replay Deep Deterministic Policy Gradient (DDPG-HER) \citep{andrychowicz2017hindsight}. The expert agent is trained by following \citet{andrychowicz2017hindsight}. For both evaluation and training, we run the expert agent for 500 and 1000 episodes, collecting the transitions as demonstrations. To avoid the imitator succeed in the task without doing any actions, we only collect the successful and non-trivial episodes from the expert. Non-trivial episodes are filtered by the following task-specific schemes:
\begin{itemize}
\item \textit{FetchReach:} An episode is considered trivial if the distance between the goal position and the initial position is smaller than 0.2.
\item \textit{FetchPush:} An episode is determined trivial if the distance between the goal position and the object position is smaller than 0.2.
\item \textit{FetchSlide:} An episode is considered trivial if the distance between the goal position and the object position is smaller than 0.1.
\item \textit{FetchPickAndPlace:} The episode is considered trivial if the distance between the goal position and the object position is smaller than 0.2.
\item \textit{HandReach:} We do not filter out trivial episodes as this task is too difficult for most of the methods.
\end{itemize}

\section{Analysis of the number of expert demonstrations}

\begin{figure}[htb]
\centering
\includegraphics[width=\linewidth]{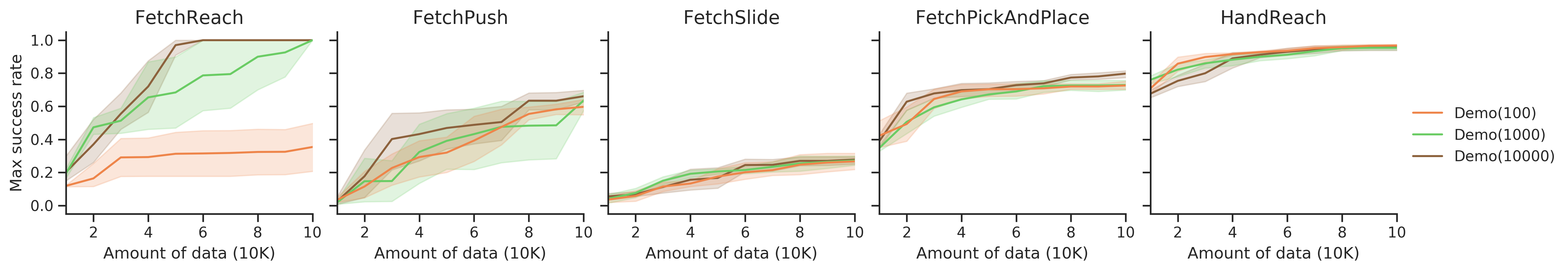}
\caption{Comparison of different number of expert demonstrations in low-dimensional observation spaces.}
\label{figure::perf_num_exper_demo}
\end{figure}

Fig.~\ref{figure::perf_num_exper_demo} illustrates the performance of \textit{Demo} with different number of expert demonstrations. \textit{Demo(100)}, \textit{Demo(1,000)}, and \textit{Demo(10,000)} correspond to the \textit{Demo} baselines with 100, 1,000, and 10,000 episodes of demonstrations, respectively. It is observed that their performance are comparable for most of the tasks except \textit{FetchReach}. In \textit{FetchReach}, the performance of \textit{Demo(100)} is significantly worse than the other two cases.  A possible explanation is that preparing a sufficiently diverse set of demonstrations in \textit{FetchReach} is relatively difficult with only 100 episodes of demonstrations.  A huge performance gap is observed when the number of episodes is increased to 1,000.  Consequently, \textit{Demo(1,000)} is selected as our \textit{Demo} baseline for the presentation of the  experimental results.  Another advantage is that \textit{Demo(1,000)} demands less memory than \textit{Demo(10,000)}.

%to the others with a large gap. The likely rationale is that it is relatively difficult to sample a diverse set of demonstration in \textit{FetchReach} with a small set of samples. Consequently, the insufficiently diverse demonstration leads to the performance drop.

%To conclude, we choose \textit{Demo(1000)} for the major experiments in Section~\ref{sec::exp} as it demands less memory than \textit{Demo(10000)} but results in a comparable performance. 

% \section{Setup of Noisy Action}
% To test the robustness of our method to noisy actions, we add noise to the actions in the training stage. Let $\hat{a}_t$ denote the predicted action by the imitator. The actual noisy action to be executed by the robot is defined as:
% \begin{equation*}
% 	\hat{a}_t := \hat{a}_t + \mathcal{N}(0, \sigma),
% \end{equation*}
% where $\sigma$ is set as $0.01$. 
% %$0.001$ in testing phase.
% Note that $\hat{a}_t$ will be clipped in the range defined by each environment.

\section{Visualization of stabilization technique}

\begin{figure}[htb]
\centering
\includegraphics[width=0.7\linewidth]{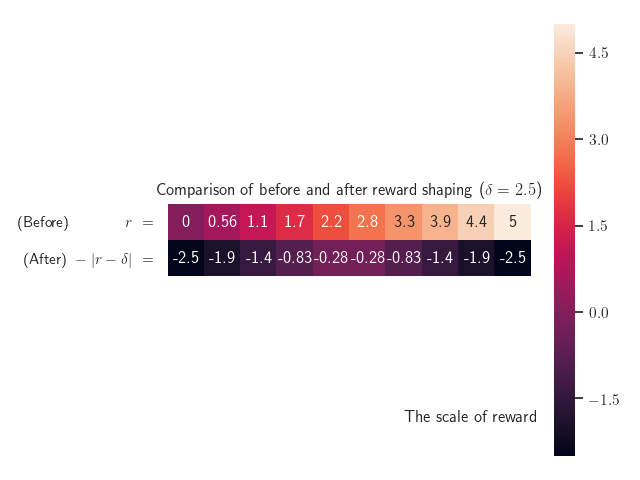}
\caption{Visualization of the stabilization technique.}
\label{figure::vis_stab}
\end{figure}

In this section, we visualize the effects of our stabilization technique with a list of rewards $r$ in Fig.~\ref{figure::vis_stab}. The rows of \textit{Before} and \textit{After} represent the rewards before and after reward shaping, respectively.
%, where the numbers among the cells in either rows denotes the corresponding value in the list.
The bar on the right-hand side indicates the scale of the reward.  It can be observed in Fig.~8 that after reward shaping, the rewards are transformed to the negative distance to the specified $\delta$ (i.e., 2.5 in this figure).  
%It can be seen that before shaping, the rewards range from 0 to 5 while after shaping the reward are transformed to negative distance to the specified $\delta$ (i.e., 2.5 for visualization).
As a result, our stabilization technique is able to encourage the DRL agent to pursue rewards close to $\delta$, where higher rewards can be received. %In sum, this visualization provides a illustrative demonstration for the effect of our stabilization technique.

\end{appendices}

\end{document}